\documentclass[letterpaper]{article} 
\usepackage{aaai2026}  
\usepackage{times}  
\usepackage{helvet}  
\usepackage{courier}  
\usepackage[hyphens]{url}  
\usepackage{graphicx} 
\usepackage{natbib}  
\usepackage{caption} 
\usepackage{algorithm}
\usepackage{algorithmic}
\usepackage{booktabs}
\usepackage{multirow}
\usepackage{array}
\usepackage{amsmath}
\usepackage{amssymb}
\usepackage{amsfonts}

\usepackage{newfloat}
\usepackage{listings}
\DeclareCaptionStyle{ruled}{labelfont=normalfont,labelsep=colon,strut=off} 
\floatstyle{ruled}
\newfloat{listing}{tb}{lst}{}
\floatname{listing}{Listing}
\title{DiGAN: Diffusion-Guided Attention Network for Early Alzheimer’s Disease Detection}
\author {
    Maxx Richard Rahman\textsuperscript{\rm 1,\rm 2},
    Mostafa Hammouda\textsuperscript{\rm 1,\rm 2},
    Wolfgang Maass\textsuperscript{\rm 1,\rm 2}
}
\affiliations {
    \textsuperscript{\rm 1}German Research Center for Artificial Intelligence (DFKI), Germany\\
    \textsuperscript{\rm 2}Saarland University, Germany\\
}

\usepackage{bibentry}

\begin{document}

\maketitle

\begin{abstract}
Early diagnosis of Alzheimer’s disease (AD) remains a major challenge due to the subtle and temporally irregular progression of structural brain changes in the prodromal stages. Existing deep learning approaches require large longitudinal datasets and often fail to model the temporal continuity and modality irregularities inherent in real-world clinical data. To address these limitations, we propose the Diffusion-Guided Attention Network (DiGAN), which integrates latent diffusion modelling with attention-guided convolutional network. The diffusion model synthesizes realistic longitudinal neuroimaging trajectories from limited training data, enriching temporal context and improving robustness to unevenly spaced visits. The attention-convolutional layer then captures discriminative structural–temporal patterns that distinguish cognitively normal subjects from those with mild cognitive impairment and subjective cognitive decline. Experiments on the ADNI dataset demonstrate that DiGAN outperforms existing state-of-the-art baselines, showing its potential for early-stage AD detection.
\end{abstract}


\section{Introduction}
Alzheimer’s disease (AD) is the most prevalent neurodegenerative disorder, accounting for 60–80\% of dementia cases worldwide~\cite{alzheimers2024facts}. It affects over 50 million individuals globally, with cases projected to triple by 2050~\cite{who2017dementia}. The resulting cognitive and functional decline imposes immense socioeconomic and healthcare burdens. Therefore, early detection of prodromal AD is important for enabling preventive interventions, guiding treatment decisions, and improving patient outcomes~\cite{dimeco2021early}. However, reliable identification of early-stage AD remains a challenge due to several factors~\cite{zhang2024recent}: (\textit{i}) subtle morphological changes in the brain precede clinical symptoms by years, (\textit{ii}) disease trajectories are highly heterogeneous across individuals, and (\textit{iii}) longitudinal neuroimaging data are limited, irregular, and often incomplete.  

Recent advances in deep learning have improved AD characterization by identifying structural and temporal biomarkers from neuroimaging data. Ensemble-based approaches like LSCP~\cite{lscp} and SUOD~\cite{suod} have improved detection robustness, while probabilistic models such as GP~\cite{gp} has been explored for anomaly detection in clinical data. Attention-based model like SACNN~\cite{sacnn} exploits structural–temporal dependencies in longitudinal data. Despite its effectiveness, existing models share key limitations: (\textit{i}) they require large amounts of longitudinal training data to generalize effectively, which is an unrealistic assumption in the healthcare domain where follow-up data are often sparse and irregular; and (\textit{ii}) they are unable to synthesize intermediate disease states reflecting continuous progression. These limitations motivate the development of models capable of generating realistic temporal trajectories and learning discriminative representations from limited and irregular clinical data.
        
To overcome these limitations, we propose the \underline{Di}ffusion-\underline{G}uided \underline{A}ttention \underline{N}etwork (DiGAN), a generative–discriminative architecture designed for early Alzheimer’s disease detection. DiGAN integrates a latent diffusion model with an attention-based convolutional network to synthesize realistic longitudinal profiles and improve temporal modeling. The diffusion component generates additional neuroimaging trajectories that mimic real-world disease progression, while the attention-convolutional encoder captures discriminative structural–temporal representations for early AD detection. The key contributions are summarized as follows:
\begin{itemize}
    \item We introduce DiGAN that fuses latent diffusion modeling with attention-based convolution for early AD detection from longitudinal neuroimaging data.
    \item We demonstrate the effectiveness of DiGAN compared to existing state-of-the-art methods on the ADNI dataset.
\end{itemize}

\begin{figure*}
    \centering
    \includegraphics[width=\linewidth]{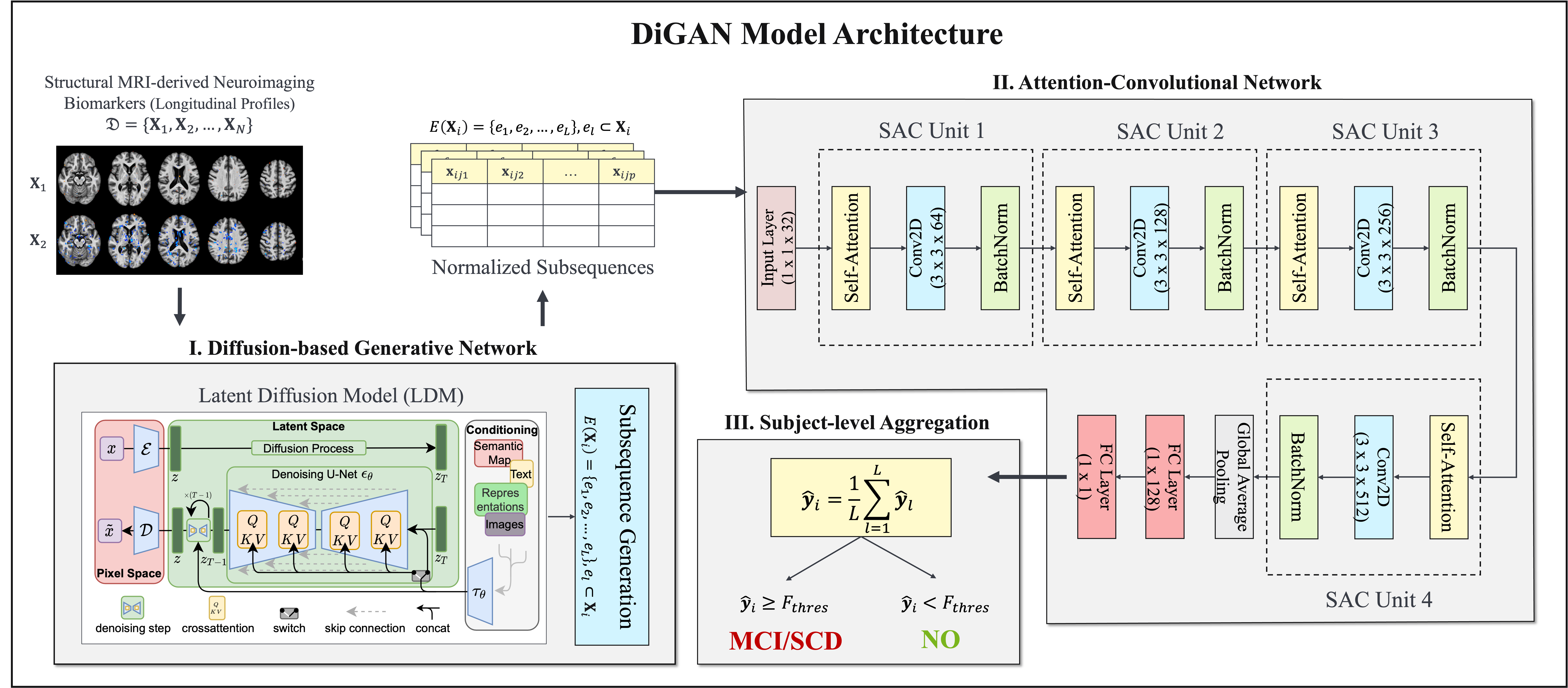}
    \caption{Overview of the DiGAN architecture comprising (i) a diffusion process for synthesizing neuroimaging profiles, (ii) an attention-convolutional network for extracting structural–temporal embeddings, and (iii) a subject-level aggregation for AD identification.}
    \label{fig:model}
\end{figure*}

\section{Preliminaries}
The objective is to detect prodromal AD by modeling temporal progression within longitudinal neuroimaging profiles. Each subject $i$ is represented by a sequence of structural MRI-derived biomarkers across baseline and follow-up visits, denoted as $\mathbf{X}_i = \{ \mathbf{x}_{i,1}, \mathbf{x}_{i,2}, \dots, \mathbf{x}_{i,n_i} \}, \quad \mathbf{x}_{i,t} \in \mathbb{R}^p$, where \(n_i\) is the number of available visits and \(p\) is the dimensionality of the biomarker feature space. Each feature vector \(\mathbf{x}_{i,t}\) includes regional brain volumes, white matter hyperintensities, perivascular spaces, etc. Each subject is assigned a diagnostic label \(y_i \), indicating cognitively normal (NO), subjective cognitive decline (SCD), mild cognitive impairment (MCI), and Alzheimer’s disease (AD). The task is to learn a discriminative function \(f_\theta: \mathbf{X}_i \mapsto y_i\) that captures both structural and temporal dependencies across visits (from baseline), thereby modeling disease progression and identifying subjects at different stages of cognitive decline.

\section{DiGAN Architecture}
As shown in Fig.~\ref{fig:model}, the DiGAN consists of three main components: i) Diffusion-based generative network, ii) Attention-convolutional network, and iii) Subject-level aggregation unit.   
   
\subsection{Diffusion-based Generative Network}
We generate neuroimaging profiles that mimic real-world progressions by following the latent diffusion framework~\cite{rombach2022high}. The forward process gradually perturbs a clean MRI-derived feature vector $\mathbf{x}_{i,t}$ into a noisy latent variable $\mathbf{z}_t$ through a sequence of Gaussian transitions: $\mathbf{z}_0 = \mathbf{x}_{i,t}, \quad 
\mathbf{z}_t = \sqrt{\alpha_t}\,\mathbf{z}_{t-1} + \sqrt{1-\alpha_t}\,\boldsymbol{\epsilon}_t, 
\quad \boldsymbol{\epsilon}_t \sim \mathcal{N}(0, I)$, where $\{\alpha_t\}_{t=1}^T$ defines the variance schedule controlling the noise magnitude. A denoising network $\mathcal{D}_\theta(\mathbf{z}_t, t)$ learns the reverse process, progressively reconstructing clean representations $\hat{\mathbf{x}}_{i,t}$ by minimizing the mean squared error between the true and predicted noise, which corresponds to maximizing a variational lower bound on $\log p(\mathbf{x})$.  

When the synthetic longitudinal profile $\hat{\mathbf{X}}_i = \{ \hat{\mathbf{x}}_{i,1}, \dots, \hat{\mathbf{x}}_{i,n_i} \}$ is generated, we apply a subsequence extraction function to handle the irregular and variable sequence lengths. The set of subsequences derived from each profile is defined as $E(\hat{\mathbf{X}}_i) = \{ e_1, e_2, \dots, e_L \}, \quad e_\ell \subset \hat{\mathbf{X}}_i$, where each subsequence $e_\ell$ corresponds to a contiguous temporal window of length $L$. All subsequences are normalized across samples to ensure consistency in feature scaling. 

\subsection{Attention-Convolutional Network}
Each normalized subsequence is fed into an attention-convolutional network composed of $m$ stacked Self-Attention Convolution (SAC) units. The purpose is to extract structural–temporal embeddings that capture both global progression trends and local variations in biomarker dynamics. In the $j^{th}$ SAC unit~\cite{sacnn}, the input sequence representation $H^{(j-1)} \in \mathbb{R}^{L \times C_{j-1}}$ (where $C_{j-1}$ denotes the number of input channels) is processed through the following operations: 

\paragraph{1. Self-Attention Layer:} Temporal dependencies across samples are modeled via a scaled dot-product attention mechanism. The attention matrix is computed as $\mathcal{A} = \mathrm{softmax}\!\left(\frac{QK^\top}{\sqrt{d_a}}\right)$, and the attention-weighted output is $H_a = \mathcal{A}V$. This allows the network to emphasize clinically relevant time points while capturing long-range dependencies across visits.
    
\paragraph{2. Convolutional Layer:} The attention-weighted representation $H_a$ is subsequently processed by a two-dimensional convolution operating over both time and feature dimensions $H^{(j)}_{\mathrm{conv}} = \sigma(W^{(j)} * H_a + b^{(j)})$, where $W^{(j)}$ and $b^{(j)}$ denote the convolutional kernels and bias, respectively, and $\sigma(\cdot)$ represents a non-linear activation function. This refines feature interactions by capturing localized structural–temporal dependencies among brain biomarkers.

\paragraph{3. Batch Normalization and Output:} To stabilize optimization and improve generalization, batch normalization is applied channel-wise $H^{(j)}_{\mathrm{BN}} = 
\frac{H^{(j)}_{\mathrm{conv}} - \mu}{\sqrt{\sigma^2 + \epsilon}} \gamma + \beta$, where $\mu$ and $\sigma^2$ denote the batch mean and variance, and $\gamma$, $\beta$ are learnable scaling and shift parameters. The resulting normalized features are propagated to the next SAC unit. After the final unit, the representation is flattened into a vector $h$, which is passed through a fully connected layer to produce a subsequence-level logit $l_{e_\ell} = w^\top h + b$, followed by a sigmoid activation $p_{e_\ell} = \sigma(l_{e_\ell})$, representing the probability that subsequence $e_\ell$ corresponds to an impaired cognitive state.
        
\subsection{Subject-Level Aggregation}
For each subject, the attention-convolutional network produces a set of subsequence-level probabilities \(\{ p_{e_\ell} \}_{\ell=1}^{L}\), where each \(p_{e_\ell}\) reflects the likelihood that the corresponding temporal window \(e_\ell \subset E(\hat{\mathbf{X}}_i)\) indicates cognitive impairment. To derive a subject-level decision, these probabilities are aggregated using a clinically motivated maximum-pooling rule $p_i = \max_{\ell} p_{e_\ell}$. A subject is classified as impaired if \(p_i \geq F_{\mathrm{thres}}\), (\(F_{\mathrm{thres}}\): predefined decision threshold). This aggregation reflects the clinical rationale that the presence of neurodegenerative evidence in any temporal segment of the longitudinal profile is sufficient to indicate elevated risk.

\subsection{Model Training and Optimization}
The diffusion model is optimized to reconstruct clean MRI-derived representations by minimizing the standard diffusion objective $\mathcal{L}_{\mathrm{diff}} = 
\mathbb{E}_{t, \mathbf{x}, \boldsymbol{\epsilon}}
\left[ \| \boldsymbol{\epsilon} - \mathcal{D}_\theta(\sqrt{\alpha_t}\,\mathbf{x} + \sqrt{1 - \alpha_t}\,\boldsymbol{\epsilon}, t) \|_2^2 \right]$. This objective corresponds to optimizing a variational lower bound on the data likelihood. The attention-convolutional network is trained using a binary cross-entropy loss augmented with an adversarial rate regularization term to mitigate class imbalance and improve specificity. Given a batch of subsequences $\{ e_b \}_{b=1}^B$ with corresponding labels $y_b \in \{0,1\}$ and predicted probabilities $p_b$, the loss function is defined as $\mathcal{L}_{\mathrm{cls}} = 
-\frac{1}{B}\sum_{b=1}^{B}
\big[y_b \log p_b + (1 - y_b)\log(1 - p_b)\big]
+ \lambda \left(\frac{\sum_{b:y_b=1} p_b}{\sum_{b:y_b=0} p_b}\right)$, where $\lambda$ controls the trade-off between classification accuracy and class-wise regularization. The second term penalizes disproportionate positive-to-negative prediction ratios, promoting higher specificity and robust discrimination under imbalanced data conditions. The overall training objective combines the diffusion reconstruction loss and the classification loss: $\mathcal{L} = \mathcal{L}_{\mathrm{diff}} + \beta\,\mathcal{L}_{\mathrm{cls}}$, where $\beta$ balances generative and discriminative learning.

\section{Experiments}
\subsection{Datasets}
\paragraph{ADNI.~\cite{petersen2010adni}}
A longitudinal study widely used for Alzheimer’s disease research. Subjects with at least two MRI sessions were included. It exhibits higher inter-subject variability and more pronounced class imbalance, providing a challenging real-world testbed for evaluating the model generalizability. Each profile comprises volumetric measures of cortical and subcortical regions, including the entorhinal cortex, parahippocampal gyrus, precuneus, posterior cingulate, and temporal lobes, as well as white matter hyperintensities (WMH) in the frontal and parietal lobes, perivascular spaces (PVS) in the basal ganglia and centrum semiovale.

\begin{table}
    \centering
    \caption{Data statistics across different diagnostic groups.}
    \label{tab:data}
    \resizebox{0.49\textwidth}{!}{
    \begin{tabular}{llccccc}
    \toprule
    \textbf{Datasets} & & \multicolumn{4}{c}{\textbf{Statistics (Male/Female)}} & \textbf{Total} \\
    \cmidrule(lr){3-6}
    & & \textbf{NO} & \textbf{MCI} & \textbf{SCD} & \textbf{AD} & \\
    \midrule
    \multirow{2}{*}{ADNI} 
    & \# subjects & 97/158 & 36/60 & - &   186/150 & \textbf{319/368} \\
    & \# visits & 158/318 & 65/117 & - &   291/236 & \textbf{514/671} \\
  \bottomrule
  \end{tabular}
    }
\end{table}
            
\subsection{Experimental Settings}
\paragraph{Tasks.}
Since this work focuses on early AD detection, we conduct classification tasks targeting the prodromal stages of impairment: NO vs. MCI and NO vs. AD. We consider \(n_i\in\{2,3,4\}\) to reflect subjects with different numbers of visits.  Evaluation metrics include accuracy, sensitivity, specificity, precision, F1-score and area under the ROC.

\paragraph{Baseline Methods.}
We compare DiGAN against state-of-the-art linear method (ALASCA~\cite{alasca}), generative models (TVAE~\cite{tvae}, AnoGAN~\cite{anogan}), probabilistic-kernel model (GP~\cite{gp}), and ensemble-based methods (LSCP~\cite{lscp}, SUOD~\cite{suod}, IsoForest~\cite{isoforest}) for longitudinal anomaly detection.

\begin{figure*}
    \centering
    \includegraphics[width=\linewidth]{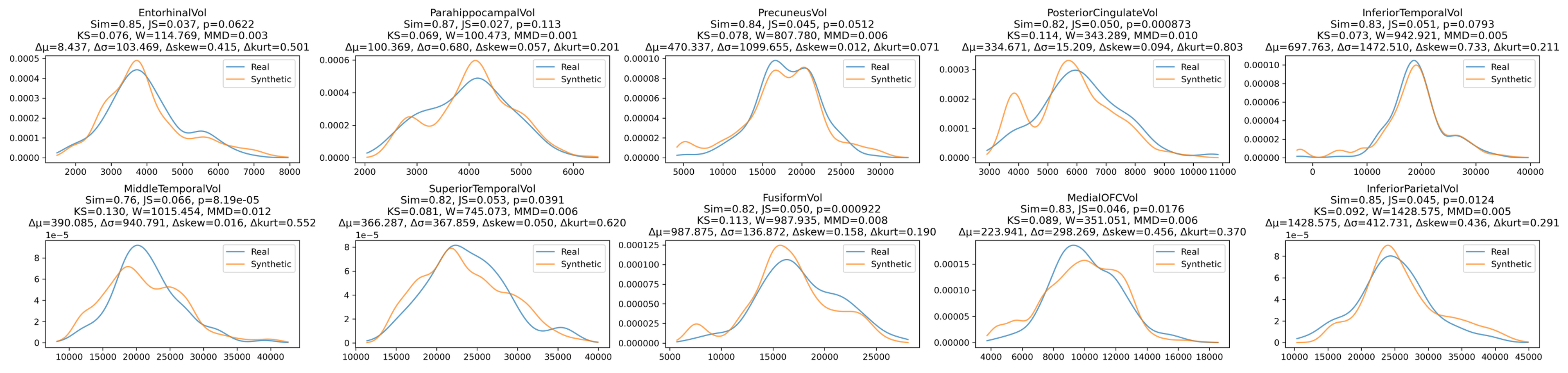}
    \caption{Comparison between real and synthetic ADNI data distributions of different parameters.}
    \label{fig:synthetic_data}
\end{figure*}

\begin{figure*}
    \centering
    \includegraphics[width=\linewidth]{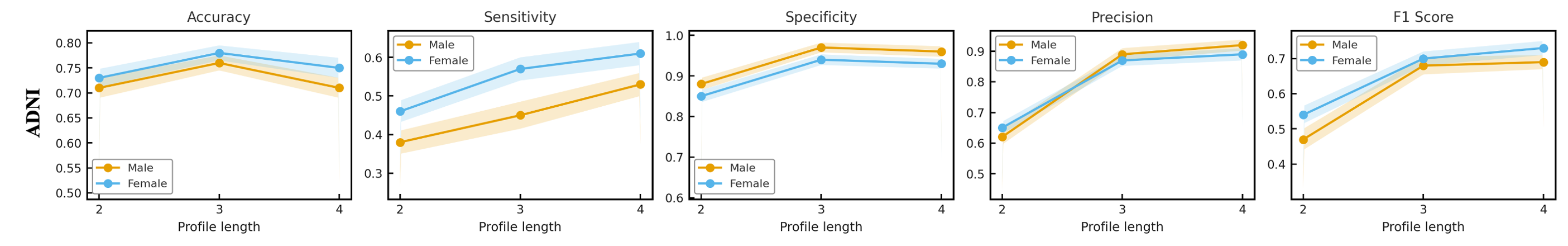}
    \caption{Performance of DiGAN for NO vs. MCI across profile lengths of 2, 3, and 4 visits.}
    \label{fig:classification_mci}
\end{figure*}

\begin{figure*}
    \centering
    \includegraphics[width=\linewidth]{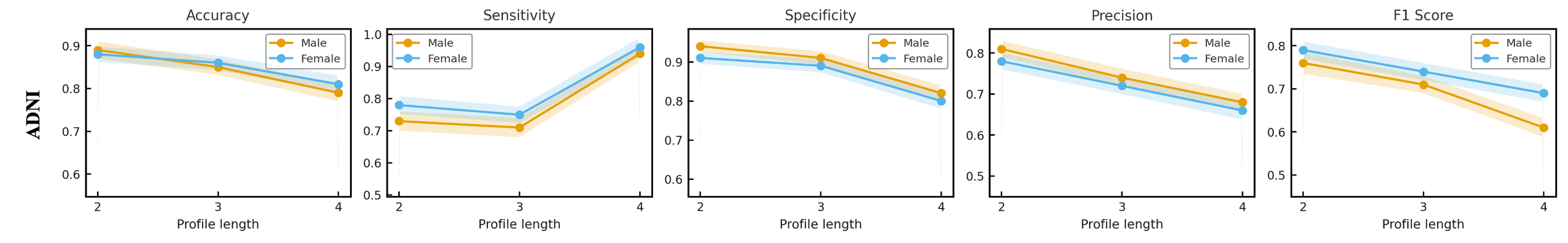}
    \caption{Performance of DiGAN for NO vs. AD across profile lengths of 2, 3, and 4 visits.}
    \label{fig:classification_scd}
\end{figure*}

\begin{figure*}
    \centering
    \includegraphics[width=\linewidth]{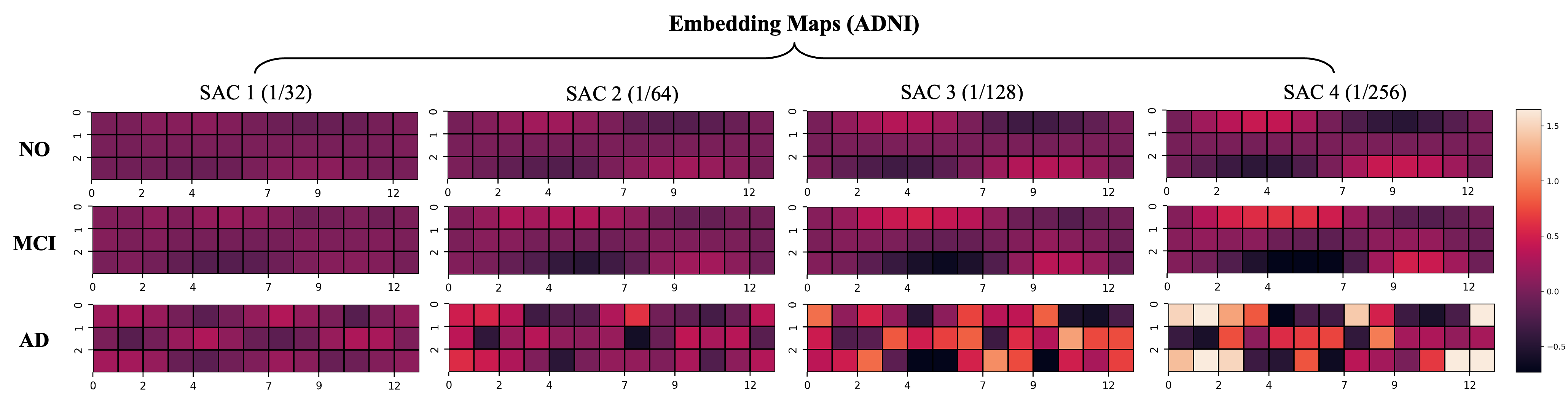}
    \caption{Embedding maps extracted from SAC units of DiGAN on the ADNI dataset. Each map shows the learned structural–temporal embeddings across subsequences.}
    \label{fig:embedding_maps}
\end{figure*}

\section{Results}
\subsection{Synthetic Dataset Evaluation}
\begin{figure}
    \centering
    \includegraphics[width=\linewidth]{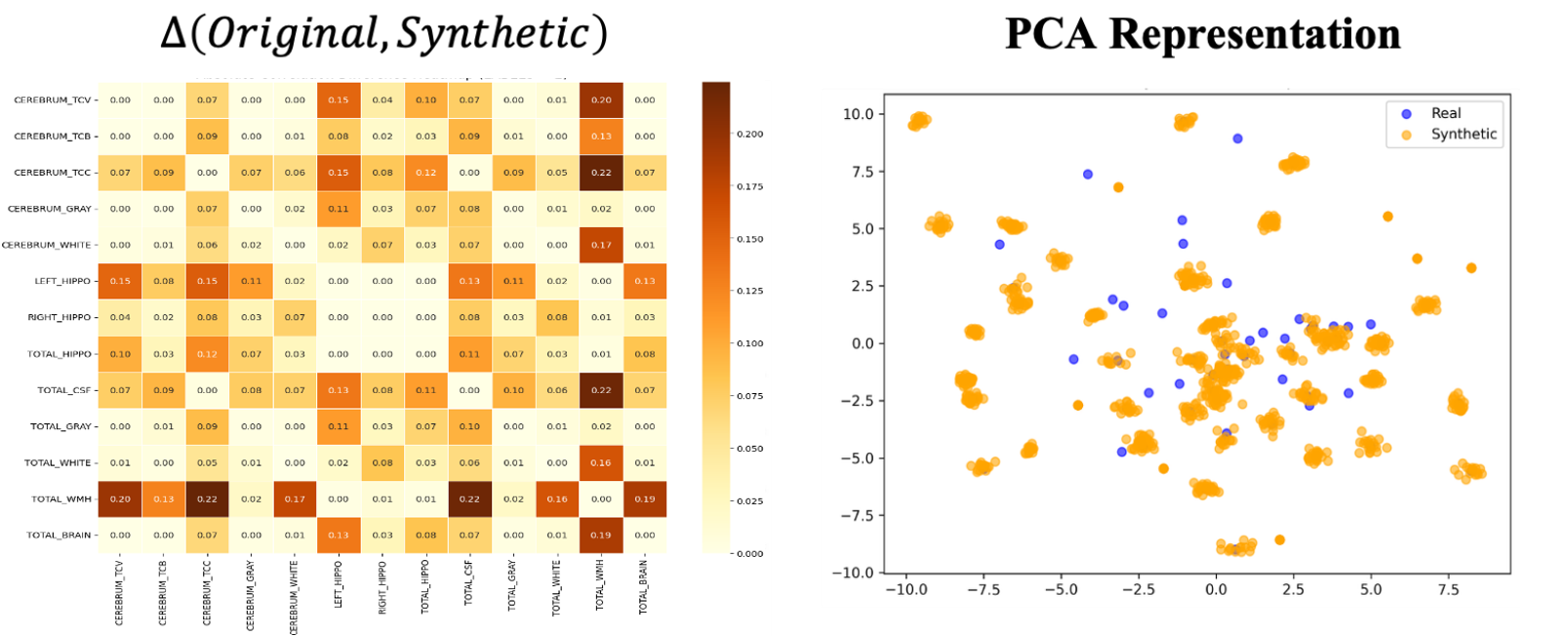}
    \caption{The differential correlation heatmap (left) and a PCA projection (right), showing substantial overlap between synthetic and real profiles.}
    \label{fig:correlation}
\end{figure}

To assess the quality of the synthetic profiles by the diffusion model, we perform a comparison analysis between the synthetic and real distributions. As shown in Fig.~\ref{fig:synthetic_data}, the probability density curves of different parameters (ADNI) show a strong overlap, with only minor deviations observed in extreme regions of the distributions. Quantitatively, similarity metrics remain consistently low, indicating high fidelity between these measurements. The differences in first- and higher-order moments (\(\Delta\mu\), \(\Delta\sigma\), skewness, and kurtosis) are within acceptable limits across most parameters, confirming that the generative process preserves anatomical variability. Fig.~\ref{fig:correlation} shows that the differential correlation heatmap and the PCA visualisation corroborate these findings by showing substantial overlap in the latent space, implying that the synthetic data accurately captures the intrinsic variance structure of the real data.

\subsection{NO vs. MCI Performance}
Fig.~\ref{fig:classification_mci} shows the classification performance of DiGAN for distinguishing cognitively normal from subjects with mild cognitive impairment. In the ADNI dataset, which contains greater inter-subject and acquisition variability, shows a more gradual improvement with increased temporal context: accuracy rises from 0.70 (2 visits) to 0.77 (3 visits), accompanied by similar gains in sensitivity and F1-score. This improvement demonstrates the model’s ability to leverage temporal continuity in longitudinal profiles for improved discrimination of subtle cognitive decline. In addition, Table~\ref{tab:delcode_adni} shows that DiGAN achieves an overall accuracy of 0.710 (ADNI), outperforming all baseline models.

\begin{table*}[htbp]
\centering
\caption{Performance comparison of different models on the ADNI dataset for NO vs. MCI.}
\label{tab:delcode_adni}
\resizebox{\textwidth}{!}{
\begin{tabular}{llcccccccc}
\toprule
\textbf{Datasets} & \textbf{Metrics} & \textbf{TVAE} & \textbf{LSCP} & \textbf{SUOD} & \textbf{GP} & \textbf{IsoForest} & \textbf{ALASCA} & \textbf{AnoGAN} & \textbf{DiGAN} \\
\midrule
\multirow{6}{*}{\textbf{ADNI}} 
& Accuracy & 0.640 $\pm$ 0.057 & 0.652 $\pm$ 0.050 & 0.630 $\pm$ 0.055 & 0.552 $\pm$ 0.120 & 0.618 $\pm$ 0.050 & 0.449 $\pm$ 0.017 & 0.618 $\pm$ 0.046 & \textbf{0.710  $\pm$ 0.042} \\
& Sensitivity & 0.485 $\pm$ 0.094 & 0.244 $\pm$ 0.108 & 0.196 $\pm$ 0.111 & 0.465 $\pm$ 0.063 & 0.220 $\pm$ 0.143 & \textbf{0.952 $\pm$ 0.041} & 0.172 $\pm$ 0.088 & 0.530  $\pm$ 0.026 \\
& Specificity & 0.771 $\pm$ 0.096 & 0.942 $\pm$ 0.001 & 0.927 $\pm$ 0.021 & 0.625 $\pm$ 0.165 & 0.958 $\pm$ 0.036 & 0.021 $\pm$ 0.036 & 0.869 $\pm$ 0.056 & \textbf{0.960  $\pm$ 0.002} \\
& AUC & 0.643 $\pm$ 0.025 & 0.572 $\pm$ 0.105 & 0.578 $\pm$ 0.103 & 0.520 $\pm$ 0.162 & 0.575 $\pm$ 0.095 & 0.495 $\pm$ 0.099 & 0.574 $\pm$ 0.096 & \textbf{0.700  $\pm$ 0.000} \\
& Precision & 0.650 $\pm$ 0.115 & 0.967 $\pm$ 0.012 & 0.906 $\pm$ 0.005 & 0.530 $\pm$ 0.148 & 0.861 $\pm$ 0.127 & 0.454 $\pm$ 0.009 & 0.889 $\pm$ 0.008 & \textbf{0.920  $\pm$ 0.071} \\
& F1-score & 0.552 $\pm$ 0.085 & 0.384 $\pm$ 0.138 & 0.318 $\pm$ 0.163 & 0.494 $\pm$ 0.100 & 0.329 $\pm$ 0.185 & 0.614 $\pm$ 0.008 & 0.287 $\pm$ 0.134 & \textbf{0.690  $\pm$ 0.016} \\

\bottomrule
\end{tabular}
}
\end{table*}

\subsection{NO vs. AD Performance}
Fig.~\ref{fig:classification_scd} shows the classification performance of DiGAN for distinguishing cognitively normal individuals from Alzheimer’s disease patients (ADNI). The DiGAN shows clear separability, achieving accuracy above 0.85 and sensitivity exceeding 0.95. As profile length increases from 2 to 4 visits, sensitivity improves consistently, substantiating the model’s ability to exploit longitudinal dynamics. Female subjects show marginally higher sensitivity and F1-scores, indicating that DiGAN effectively captures nuanced temporal biomarkers associated with early neurodegeneration.
        
\subsection{Precision-Recall and ROC Analysis}

\begin{figure}
    \centering
    \includegraphics[width=\linewidth]{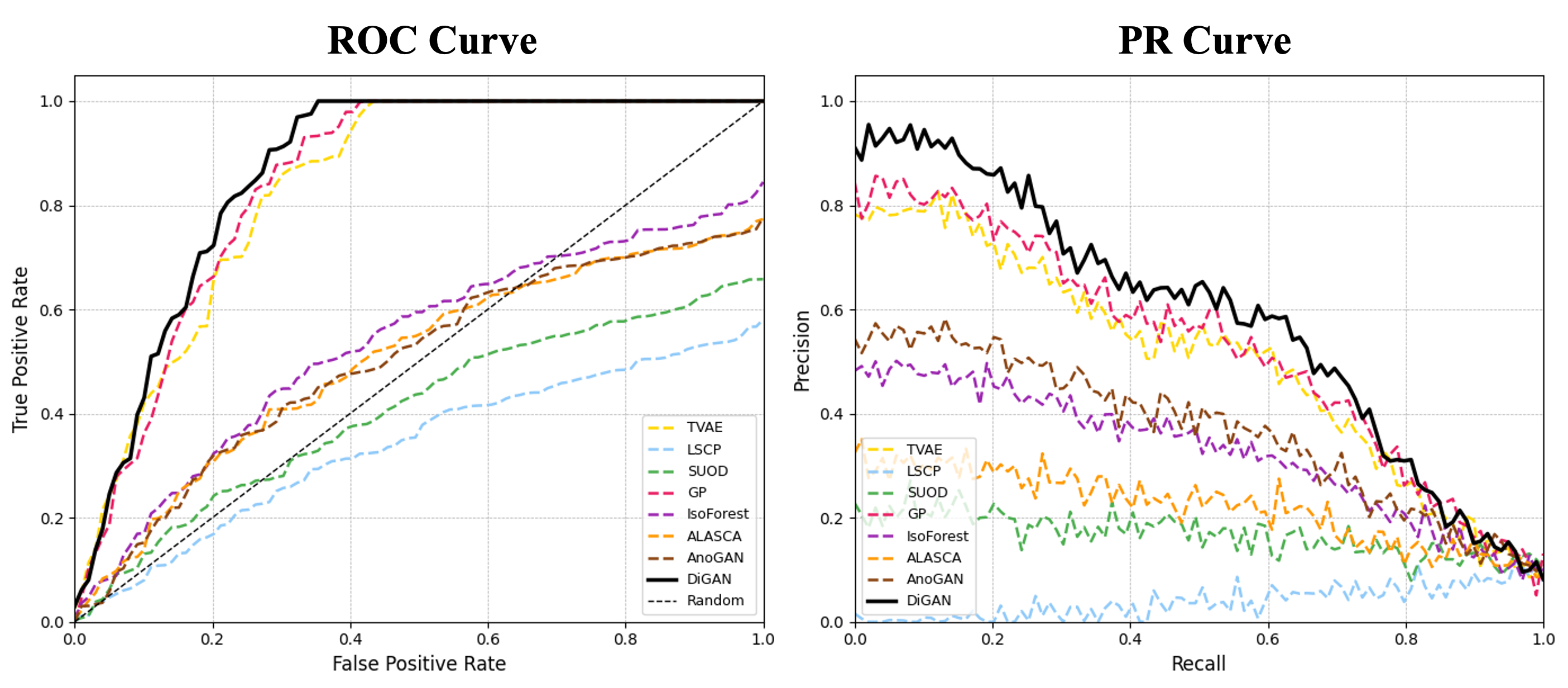}
    \caption{ROC and PR curves comparing the DiGAN performance for NO vs. MCI.}
    \label{fig:roc_curve}
\end{figure}

Fig.~\ref{fig:roc_curve} presents the ROC and precision–recall (PR) curves comparing DiGAN with baseline methods for NO vs. MCI (male subjects). DiGAN consistently outperforms all state-of-the-art models, with the ROC curve indicating good discrimination between classes even under high specificity constraints. In the PR analysis, DiGAN achieves substantially higher average precision across all recall levels, maintaining stable performance in high-recall regions where baselines degrade rapidly. This robustness indicates that the diffusion-guided representation effectively captures subtle structural and temporal deviations associated with early cognitive decline. Among baselines, GP and TVAE perform moderately well but exhibit trade-offs between precision and recall. DiGAN’s advantage arises from the integration of the diffusion process, which enriches the latent space by reconstructing intermediate neuroanatomical trajectories, and the attention mechanism, which selectively amplifies clinically relevant temporal dependencies. 

\subsection{Embedding Maps from SAC Units}
To better understand the representational dynamics within the DiGAN, we visualize the intermediate embedding maps generated by SAC units on the ADNI dataset, as shown in Fig.~\ref{fig:embedding_maps}. The early SAC units (SAC 1 and SAC 2) exhibit diffuse and low-intensity activation patterns, indicating that the network captures broad, non-discriminative features common to both cognitively normal and mildly impaired subjects. In contrast, deeper SAC units (SAC 3 and SAC 4) show increasingly structured and high-intensity activations, corresponding to more localized and discriminative representations of disease-related changes. While clear separation emerges in later layers for subjects with AD, the overlap between NO and MCI profiles remains evident, reflecting the subtle and heterogeneous nature of early neurodegenerative alterations. This progressive refinement of embeddings highlights how DiGAN incrementally separates global structural patterns from localized disease-specific signals, offering interpretability and explaining why distinguishing NO from MCI is inherently more challenging than differentiating NO from AD.
    
\section{Conclusion}
In this paper, we present DiGAN, which integrates a latent diffusion model with an attention-based convolutional encoder for early detection of AD. By generating realistic longitudinal neuroimaging trajectories from limited training data, DiGAN addresses the scarcity and irregularity of clinical follow-ups and improves temporal continuity in disease modeling. Experiments on the ADNI dataset show that DiGAN consistently outperforms existing baselines, while providing interpretable embedding and attention maps that align with known neuroanatomical markers of early AD progression.

\section{Acknowledgments}
We are thankful to Gabriel Ziegler (DZNE) and team for his valuable feedback throughout this study.

Data used in the preparation of this article were obtained from the Alzheimer’s Disease Neuroimaging Initiative (ADNI) database (adni.loni.usc.edu). The ADNI was launched in 2003 as a public-private partnership, led by Principal Investigator Michael W. Weiner, MD. The primary goal of ADNI has been to test whether serial magnetic resonance imaging (MRI), positron emission tomography (PET), other biological markers, and clinical and neuropsychological assessment can be combined to measure the progression of mild cognitive impairment (MCI) and early Alzheimer’s disease (AD).

\bibliography{aaai2026}

@article{petersen2010adni,
  title     = {Alzheimer’s Disease Neuroimaging Initiative (ADNI): Clinical characterization},
  author    = {Petersen, Ronald C. and Aisen, Paul S. and Beckett, Laurel A. and Donohue, Michael C. and Gamst, Anthony and Harvey, Danielle J. and Jack, Clifford R. and Jagust, William J. and Shaw, Leslie M. and Toga, Arthur W. and Trojanowski, John Q. and Weiner, Michael W.},
  journal   = {Alzheimer’s \& Dementia},
  volume    = {6},
  number    = {3},
  pages     = {239--246},
  year      = {2010},
  publisher = {Elsevier},
  doi       = {10.1016/j.jalz.2010.03.008}
}

@inproceedings{suod,
    title = {SUOD: Accelerating Large-scale Unsupervised Heterogeneous Outlier Detection},
    author = {Zhao, Yue and Hu, Xiyang and Cheng, Cheng and Wan, Changlin and Wang, Wen and Yang, Jianing and Bai, Haoping and Li, Zheng and Xiao, Cao and Wang, Yunlong and Qiao, Zhi and Sun, J. and Akoglu, Leman},
    booktitle = {Proceedings of Machine Learning and Systems},
    year = {2021}
}

@inproceedings{lscp,
    title={{LSCP:} Locally Selective Combination in Parallel Outlier Ensembles},
    author={Zhao, Yue and Nasrullah, Zain and Hryniewicki, Maciej K and Li, Zheng},
    booktitle={Proceedings of the 2019 {SIAM} International Conference on Data Mining, {SDM} 2019},
    pages={585--593},
    month = {May},
    year={2019},
    url={https://doi.org/10.1137/1.9781611975673.66},
    doi={10.1137/1.9781611975673.66}
}

@inproceedings{anogan,
    address = {Cham},
    author = {Schlegl, Thomas and Seeb{\"o}ck, Philipp and Waldstein, Sebastian M. and Schmidt-Erfurth, Ursula and Langs, Georg},
    booktitle = {Information Processing in Medical Imaging},
    pages = {146--157},
    publisher = {Springer International Publishing},
    title = {Unsupervised Anomaly Detection with Generative Adversarial Networks to Guide Marker Discovery},
    year = {2017}
}

@inproceedings{isoforest,
  title     = {Isolation Forest},
  author    = {Liu, Fei Tony and Ting, Kai Ming and Zhou, Zhi-Hua},
  booktitle = {2008 Eighth IEEE International Conference on Data Mining (ICDM)},
  pages     = {413--422},
  year      = {2008},
  publisher = {IEEE},
  doi       = {10.1109/ICDM.2008.17},
  url       = {https://dl.acm.org/doi/10.1109/icdm.2008.17}
}

@article{tvae,
  title     = {Rethinking Robust Multivariate Time Series Anomaly Detection: A Hierarchical Spatio-Temporal Variational Perspective},
  author    = {Zhang, Xinyu and Xu, Shun and Chen, Haifeng and Chen, Zhiqiang and Zhuang, Fuzhen and Xiong, Hui and Yu, Dejing},
  journal   = {IEEE Transactions on Knowledge and Data Engineering},
  year      = {2024},
  publisher = {IEEE},
  doi       = {10.1109/TKDE.2024.3386864},
  url       = {https://doi.org/10.1109/TKDE.2024.3386864}
}

@article{gp,
  title     = {STGP: Spatio-temporal Gaussian process models for longitudinal neuroimaging data},
  author    = {Hyun, Jin Woo and Li, Ying and Huang, Chong and Styner, Martin and Lin, Weili and Zhu, Hongtu and Alzheimer's Disease Neuroimaging Initiative},
  journal   = {NeuroImage},
  volume    = {134},
  pages     = {550--562},
  year      = {2016},
  publisher = {Elsevier},
  doi       = {10.1016/j.neuroimage.2016.04.044},
  url       = {https://doi.org/10.1016/j.neuroimage.2016.04.044}
}

@article{alasca,
  title={ALaSCA: a computational platform for quantifying the effect of proteins using Pearlian causal inference, with an example application in Alzheimer’s disease},
  author={Truter, N. and Jansen van Rensburg, Z. and Oudrhiri, R. and Van Niekerk, D. D. and Loos, B. and Singh, R. and Louw, C.},
  journal={bioRxiv},
  year={2022},
  pages={2022--10},
  publisher={Cold Spring Harbor Laboratory}
}

@inproceedings{rombach2022high,
  title     = {High-Resolution Image Synthesis with Latent Diffusion Models},
  author    = {Rombach, Robin and Blattmann, Andreas and Lorenz, Dominik and Esser, Patrick and Ommer, Bj{\"o}rn},
  booktitle = {Proceedings of the IEEE/CVF Conference on Computer Vision and Pattern Recognition (CVPR)},
  pages     = {10684--10695},
  year      = {2022},
  doi       = {10.1109/CVPR52688.2022.01042},
  url       = {https://doi.org/10.1109/CVPR52688.2022.01042}
}

@inproceedings{sacnn,
  title     = {SACNN: Self Attention‐based Convolutional Neural Network for Fraudulent Behaviour Detection in Sports},
  author    = {Rahman, Maxx Richard and Abdel Khaliq, Lotfy and Piper, Thomas and Geyer, Hans and Equey, Tristan and Baume, Norbert and Aikin, Reid and Maass, Wolfgang},
  booktitle = {Proceedings of the Thirty-Third International Joint Conference on Artificial Intelligence (IJCAI-24)},
  pages     = {6017--6025},
  year      = {2024},
  publisher = {International Joint Conferences on Artificial Intelligence},
  doi       = {10.24963/ijcai.2024/665},
  url       = {https://doi.org/10.24963/ijcai.2024/665}
}

@article{alzheimers2024facts,
  title     = {2024 Alzheimer’s disease facts and figures},
  journal   = {Alzheimer’s \& Dementia: The Journal of the Alzheimer’s Association},
  author    = {Alzheimer},
  volume    = {20},
  number    = {5},
  pages     = {3708--3821},
  year      = {2024},
  publisher = {Wiley},
  doi       = {10.1002/alz.13809},
  url       = {https://doi.org/10.1002/alz.13809}
}

@incollection{dimeco2021early,
  title     = {Early detection and personalized medicine: Future strategies against Alzheimer's disease},
  author    = {Di Meco, Antonio and Vassar, Robert},
  booktitle = {Progress in Molecular Biology and Translational Science},
  volume    = {177},
  pages     = {157--173},
  year      = {2021},
  publisher = {Elsevier},
  doi       = {10.1016/bs.pmbts.2020.10.002},
  url       = {https://doi.org/10.1016/bs.pmbts.2020.10.002}
}

@article{zhang2024recent,
  title     = {Recent advances in Alzheimer’s disease: mechanisms, clinical trials and new drug development strategies},
  author    = {Zhang, Jie and Zhang, Yufei and Wang, Jian and others},
  journal   = {Signal Transduction and Targeted Therapy},
  volume    = {9},
  number    = {1},
  pages     = {211},
  year      = {2024},
  publisher = {Springer Nature},
  doi       = {10.1038/s41392-024-01911-3},
  url       = {https://doi.org/10.1038/s41392-024-01911-3}
}

@misc{who2017dementia,
  author       = {WHO},
  title        = {Dementia: number of people affected to triple in next 30 years},
  month        = dec,
  year         = {2017},
  note         = {News release. Geneva}
}

\end{document}